\documentclass[conference]{IEEEtran}
\usepackage{times}

\usepackage[numbers]{natbib}
\usepackage{multicol}
\usepackage[bookmarks=true]{hyperref}

\usepackage[utf8]{inputenc} % allow utf-8 input
\usepackage[T1]{fontenc}    % use 8-bit T1 fonts
\usepackage{hyperref}       % hyperlinks
\usepackage{url}            % simple URL typesetting
\usepackage{booktabs}       % professional-quality tables
\usepackage{amsfonts}       % blackboard math symbols
\usepackage{nicefrac}       % compact symbols for 1/2, etc.
\usepackage{microtype}      % microtypography
\usepackage{graphicx}
\usepackage{wrapfig}
\usepackage{amsmath}

\begin{document}

\title{Value-Based Reinforcement Learning for Continuous Control Robotic Manipulation in Multi-Task Sparse Reward Settings}

% You will get a Paper-ID when submitting a pdf file to the conference system
\author{\authorblockN{Sreehari Rammohan\authorrefmark{1},
Shangqun Yu\authorrefmark{1},
Bowen He\authorrefmark{1}, 
Eric Hsiung\authorrefmark{1},
Eric Rosen,
Stefanie Tellex, George Konidaris}\authorblockA{\authorrefmark{1} These authors contributed equally.\\
Brown University\\}}

%\author{\authorblockN{Michael Shell}
%\authorblockA{School of Electrical and\\Computer Engineering\\
%Georgia Institute of Technology\\
%Atlanta, Georgia 30332--0250\\
%Email: mshell@ece.gatech.edu}
%\and
%\authorblockA{Twentieth Century Fox\\
%\and
%\authorblockN{James Kirk\\ and Montgomery Scott}
%\authorblockA{Starfleet Academy\\
%San Francisco, California 96678-2391\\

% avoiding spaces at the end of the author lines is not a problem with
% conference papers because we don't use \thanks or \IEEEmembership

%Homer Simpson\authorrefmark{2},
%Montgomery Scott\authorrefmark{3} and
%Eldon Tyrell\authorrefmark{4}}
%Telephone: (800) 555--1212, Fax: (888) 555--1212}
%\authorblockA{\authorrefmark{4}Tyrell Inc., 123 Replicant Street, Los Angeles, California 90210--4321}}

\maketitle

\begin{abstract}
Learning continuous control in high-dimensional sparse reward settings, such as robotic manipulation, is a challenging problem due to the number of samples often required to obtain accurate optimal value and policy estimates. While many deep reinforcement learning methods have aimed at improving sample efficiency through replay or improved exploration techniques, state of the art actor-critic and policy gradient methods still suffer from the hard exploration problem in sparse reward settings. Motivated by recent successes of value-based methods for approximating state-action values, like RBF-DQN, we explore the potential of value-based reinforcement learning for learning continuous robotic manipulation tasks in multi-task sparse reward settings. On robotic manipulation tasks, we empirically show RBF-DQN converges faster than current state of the art algorithms such as TD3, SAC, and PPO. We also perform ablation studies with RBF-DQN and have shown that some enhancement techniques for vanilla Deep Q learning such as Hindsight Experience Replay (HER) and Prioritized Experience Replay (PER) can also be applied to RBF-DQN. Our experimental analysis suggests that value-based approaches may be more sensitive to data augmentation and replay buffer sample techniques than policy-gradient methods, and that the benefits of these methods for robot manipulation are heavily dependent on the transition dynamics of generated subgoal states. 
\end{abstract}

\IEEEpeerreviewmaketitle

\section{Introduction} % Establish the problem
%Great progress has been made in the field of reinforcement learning (RL)~\cite{sutton98} by combining RL with deep learning. Deep reinforcement learning agents can now successfully execute complex tasks in large, high-dimensional state spaces such as achieving human level game play on Atari 2600 games \cite{Mnih2015HumanlevelCT}, or beating professional Go players \cite{Silver2017MasteringTG}. However, robotic manipulation using deep RL methods remains a challenging problem \cite{kilinc2020reinforcement}, because the combination of continuous high-dimensional state and action spaces coupled with sparse rewards presents a difficult exploration problem and subsequently makes it difficult to learn in a sample-efficient manner. For example, lifting a block first requires finding an effective, task-oriented grasp in order to succeed.
% What are the limits of existing work?

Current RL algorithms aimed at robot manipulation are either policy gradient methods, such as PPO~\cite{schulman2017proximal}, or actor-critic methods, such as DDPG~\cite{lillicrap2019continuous}, TD3~\cite{fujimoto2018addressing}, or SAC~\cite{haarnoja2018soft}. These methods have a stable learning process because they directly optimize policy parameters based on expected return, but still suffer from sample inefficient function approximation when compared to value-based optimization approaches. While value-based approaches were previously either limited by their function approximation capability or could only work in discrete action spaces, recent works \cite{asadi2020} have developed novel neural network architectures that enable general function approximation of continuous state-action value functions only using the Bellman error \cite{asadi2020}.

%These methods incorporate various exploration-exploitation techniques, but still suffer from sample inefficiency when approximating the optimal value or policy when compared to value-based approaches. While value-based approaches were previously either limited by their function approximation capability or could only work in discrete action spaces, recent works \cite{asadi2020} have developed novel neural network architectures that enable general function approximation of continuous state-action value functions only using the Bellman error \cite{asadi2020}.
%Furthermore, choosing optimal actions in any given state hinges on accurate mappings from states to actions, whether via policy or value-estimates.

In this paper we evaluate RBF-DQN~\cite{asadi2020}, a action value-based method inspired by Q-networks, to complete multiple RLBench \cite{james2019} tasks in continuous state and action spaces. RLBench provides a challenging testbed for evaluating reinforcement learning algorithms in multi-task robot settings because it only provides sparse-rewards for completing long-horizon tasks. We conduct experiments in 5 different tasks (Fetch Reach, Button Push, Toilet Down, Open Drawer, and Pick and Place), and investigate how value-based approaches (like RBF-DQN) compare to state-of-the-art actor-critic methods (like T3D, PPO, and SAC) and how their performance is impacted by typical data augmentation and replay buffer sampling techniques. To the best of our knowledge, this is the first comparison of value-based approaches against actor-critic methods for continuous robot control in sparse reward multi-task settings.

\begin{figure}
    \centering
    \includegraphics[scale=0.4]{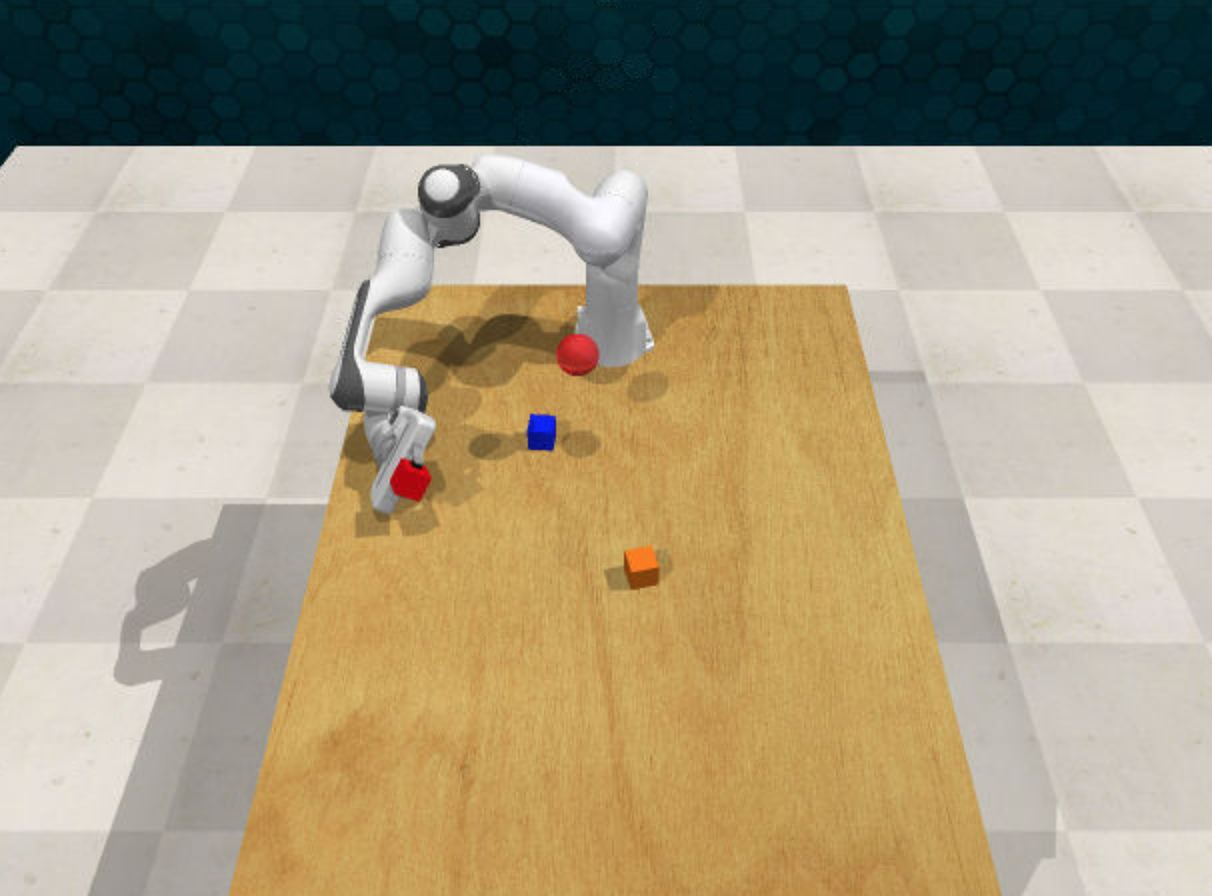}
    \caption{\textit{Pick and Lift} RLBench Task. The agent is tasked with picking up the red block and moving it to the point in space represented by the red sphere.}
    \label{fig:my_label}
\end{figure}

%EH: Not necessary in intro: Furthermore, as opposed to explicit reward engineering, our agent is only exposed to a sparse reward signal, receiving a reward of 1 for successfully completing the task and 0 in all other transitions.

%\ernote{I'd suggest reading the heilmeier catechisms, and make sure your introduction hits all the points}

\section{Background and related work}
Reinforcement learning is the study of maximizing an agent's long term discounted reward through interactions with an environment \cite{sutton98}.
It is commonly modeled as a Markov Decision Process (MDP) ~\cite{Puterman94}, defined by the tuple $ \langle \mathcal{S}, \mathcal{A}, T, R, \gamma \rangle $. In robotic manipulation domains, $\mathcal{S}$ denotes the continuous state space, $\mathcal{A}$ represents the continuous action space, $T:  \mathcal{S} \times \mathcal{A} \times \mathcal{S} \rightarrow [0, 1] $ is the transition model, and $R: \mathcal{S} \times \mathcal{A} \rightarrow \mathbb{R}$ is the reward model.
The discount factor, $\gamma \in [0,1)$, 
determines the importance of immediate rewards compared with future rewards. 
The action-value function $Q^{\pi}(s,a)$, with $s\in\mathcal{S}$ and $a\in\mathcal{A}$, is defined as the maximum expected return achievable by following a particular policy $\pi: \mathcal{S} \rightarrow \mathcal{A}$ \cite{DBLP:journals/corr/MnihKSGAWR13}, after seeing some state $s$ and then taking some action $a$. The optimal action-value function
$Q^{*}(s,a) = \text{max}_{\pi} \mathop{\mathbb{E}}[R_t | s_t = s, a_t = a, \pi]$ corresponds with the optimal policy.
The optimal action-value function follows an important identity which is known as the Bellman equation ~\cite{Bellman1952OnTT}:
\begin{equation}
Q^{*}(s, a) = R(s, a) + \gamma  \int_{s'} T(s, a, s') \max_{a'} Q^{*}(s',a') \,ds' 
\end{equation}
\subsection{Q-Learning}
When the reward function and transition function are known, then the optimal $Q^*$ can be easily found by using standard dynamic programming algorithms such as value iteration. However if the model dynamics are not known, RL algorithms will need to find $Q^*$ by interacting with the environment without learning an explicit model. One notable example of these model-free algorithms is Q-learning ~\cite{qlearning}, which approximates $Q^*(s,a)$ using an estimator $\hat{Q}(s,a;\theta)$ that depends on parameters $\theta$. The parameters $\theta$ can be updated iteratively through gradient descent using the $Q$ estimates (often stabilized using a target network, such as in DQN):

\begin{equation} \label{eq:2}
\begin{aligned}
 \theta \leftarrow \theta + \alpha \delta \nabla_\theta \hat{Q}(s,a;\theta), \\
\textrm{where } \delta := r + \gamma \max_{a' \in \mathcal{A}} \hat{Q}(s', a'; \theta) - \hat{Q}(s, a; \theta)
\end{aligned}
\end{equation}

\subsection{RBF-DQN}
Introduced by \citet{asadi2020}, RBF-DQN is a value-based method that can efficiently approximate $Q^*(s,a)$ using a set of radial basis functions, and simultaneously approximates the action maximizing $Q$-value with bounded error. Specifically, RBF-DQN approximates $Q^*(s,a)$ by optimizing centroid locations $a_i(s;\theta)$ and centroid values $v_i(s;\theta)$ as functions of state $s$ and parameters $\theta$ and $\beta$ with the following equation:
\cite{asadi2020}:
\begin{equation} 
 \hat{Q}_\beta(s,a;\theta) := \frac{\sum^{N}_{i=1}e^{-\beta \lVert a - a_i(s;\theta) \rVert }v_i(s;\theta)}{\sum^{N}_{i=1} e^{-\beta \lVert a - a_i(s;\theta) \rVert}}
\end{equation}

During training, both the centroid locations $a_i(s; \theta)$ and state-dependent centroid values $v_i(s;\theta)$ are learned, these are then used during forward propagation to form the Q function output \cite{asadi2020}. In multi-dimensional action spaces, the temperature parameter $\beta$ can be tuned to ensure an upper bound error \cite{asadi2020}:
\begin{equation}
\text{max}_{a\in\mathcal{A}}\hat{Q}_\beta(s,a;\theta)-\text{max}_{i\in[1,N]}\hat{Q}_\beta(s,a;\theta)\leq\mathcal{O}(e^{-\beta})
\end{equation}

The key to why RBF-DQN is so powerful as a value-based method in continuous action spaces is due to its action-maximization property as well as the fact that it is a universal function approximator \cite{asadi2020}. In Q learning, the update rule (\ref{eq:2}) relies on finding $\max_{a' \in \mathcal{A}} \hat{Q}(s', a'; \theta)$. This is prohibitively expensive in continuous action spaces, due to a nearly infinite search space, and employing tricks like discretizing the action space may produce sub-optimal solutions. The action maximization property of RBF-DQN however, guarantees that all critical points of $\hat{Q}_\beta$ can be well-approximated by a centroid location $a_i$ \cite{asadi2020}. This makes action-maximization as simple as searching over all $N$ centroids $\max_{i \in [1, N]} \hat{Q}_\beta(s, a_i; \theta)$ where $a_i$ represents a centroid location.
\subsection{HER and PER}
Most robotic manipulation tasks are under the sparse reward setting, which makes training an RL agent extremely challenging due ineffective exploration, leading to high sample inefficiency. Hindsight Experience Replay (HER) \cite{andry2017} and Prioritized Experience Replay (PER) \cite{schaul2015} are two methods which can be used to improve the sample efficiency of previously experienced states. As agents train, $(s,a,s',r,g)$ transition tuples are collected and stored in a replay buffer $\mathcal{R}$, where $s,s'\in\mathcal{S}, a\in\mathcal{A}, r=R(s,a,s')$, and $g$ is the goal. These transitions often come from trajectories generated by the agent's policy during each episode, and they are stored in the replay buffer as a dataset of samples to train with.

In HER, after each training episode, both the original goal $g$ and potentially multiple hindsight goals $g'$ are selected from the current trajectory according to a goal selection strategy, and these are stored in the replay buffer.

In PER \cite{schaul2015}, transitions are sampled from the replay buffer weighted by their TD or Bellman error, rather than being sampled uniformly. This conceptually means that the agent prioritizes transitions in the replay buffer which it finds surprising or unexpected.

In \cite{schaul2015}, the probability $P(i)$ of sampling a transition $i$ is based on the priority of the transition $p_i=|\delta_i| + \epsilon$:

\begin{equation}
\begin{aligned}
P(i) = \frac{p_{i}^{\alpha}}{\sum_k p_{k}^{\alpha}} \\
\end{aligned}
\end{equation}
The hyperparameter $\alpha$ determines the degree to which prioritization is used.
\section{Technical Approach}

Our aim is to demonstrate RBF-DQN's efficacy on robot manipulation tasks. We applied RBF-DQN to various simulated robotic manipulation tasks under sparse rewards to investigate RBF-DQN's performance on these tasks. We also investigated how combining HER and PER with RBF-DQN impacted performance.
\subsection{RLBench}

RLBench is a robot learning simulator with many realistic \& challenging tasks involving a Franka Panda Arm, such as \textit{Fetch Reach}, \textit{Open Door}, and \textit{Close Toilet} \cite{james2019}. One aim of RLBench is to provide a standardized suite of tasks for benchmarking performance of RL strategies. %We aim to combine several state of the art RL strategies to train agents to perform a series of tasks in RLBench, providing benchmark performance scores which we believe will be helpful for the RL community.

\subsection{Goal Selection and Detection}
For our robotic manipulation tasks, we utilized two hindsight goal selection strategies for use with HER. A simple hindsight goal selection strategy, known as \emph{final}, passes the last state of a trajectory into a function $\phi$ which maps states to goals. We also considered a strategy called \emph{future}, which considers states later on in the trajectory relative to a given timestep $t$ as goals.

In our ablation studies with RBF-DQN involving HER, we use the \emph{final} and \emph{future} strategy where we not only use the final state as a hindsight goal, but also $k=4$ \emph{future} states later on in the trajectory relative to a given timestep $t$.

The specifics behind $\phi$ depend largely on the manipulation task being performed. For \textit{Fetch Reach}, $\phi$ takes the state as input and returns the $(x, y, z)$ position of the end effector, but for a task like \textit{Open Drawer}, $\phi$ returns the state of the prismatic joint representing how open or closed the drawer is.

Finally, we determine whether a goal was achieved by checking if the norm of the achieved $\phi(s)$ and desired goal $g$ is less than some arbitrary $\epsilon$ (which for our experiments has been set to $1e^{-2}$): $\|\phi(s)-g\|\leq \epsilon$.

\section{Experiments}
\begin{figure*}
    \centering
    \includegraphics[width=0.20\textwidth]{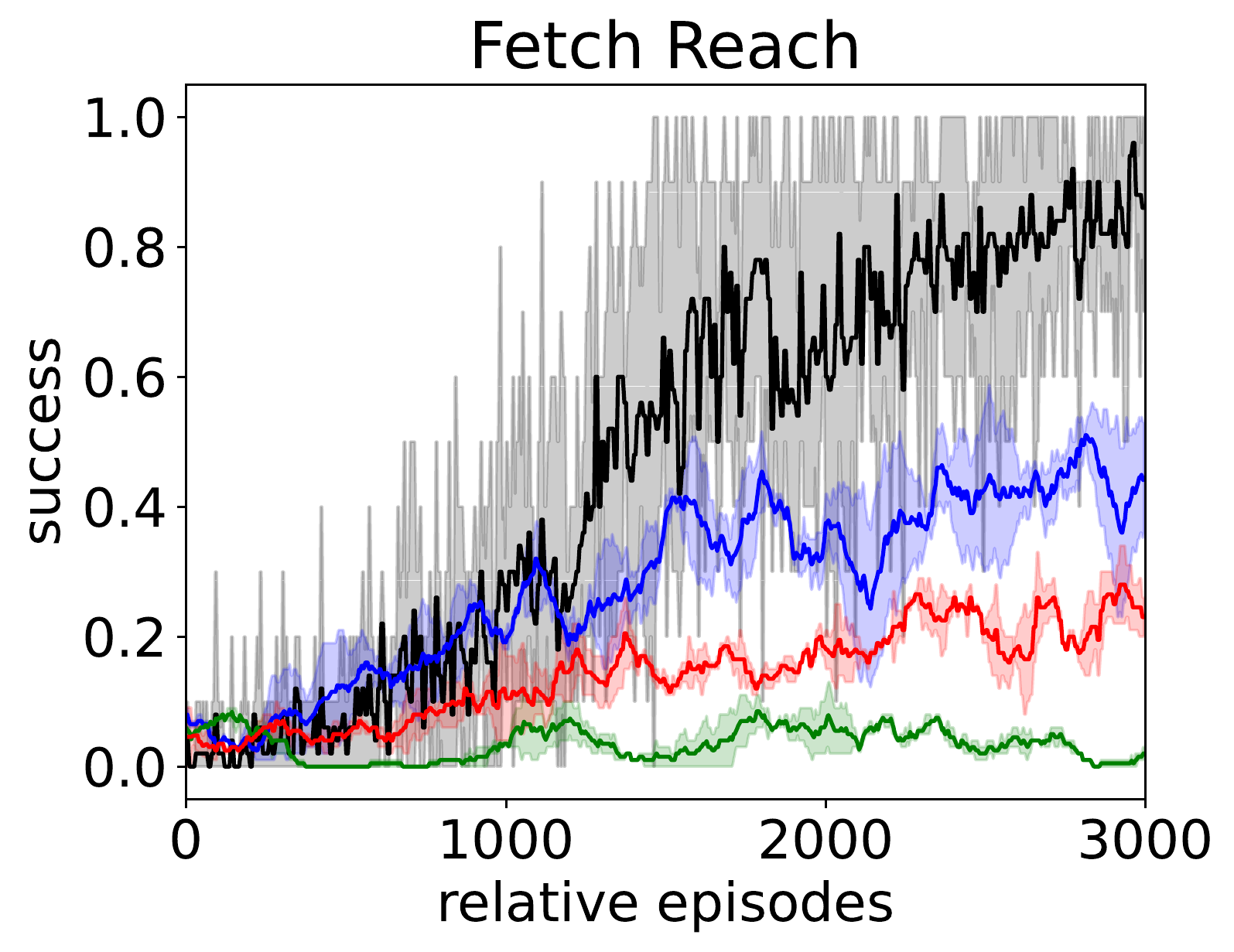}\hfill
    \includegraphics[width=0.20\textwidth]{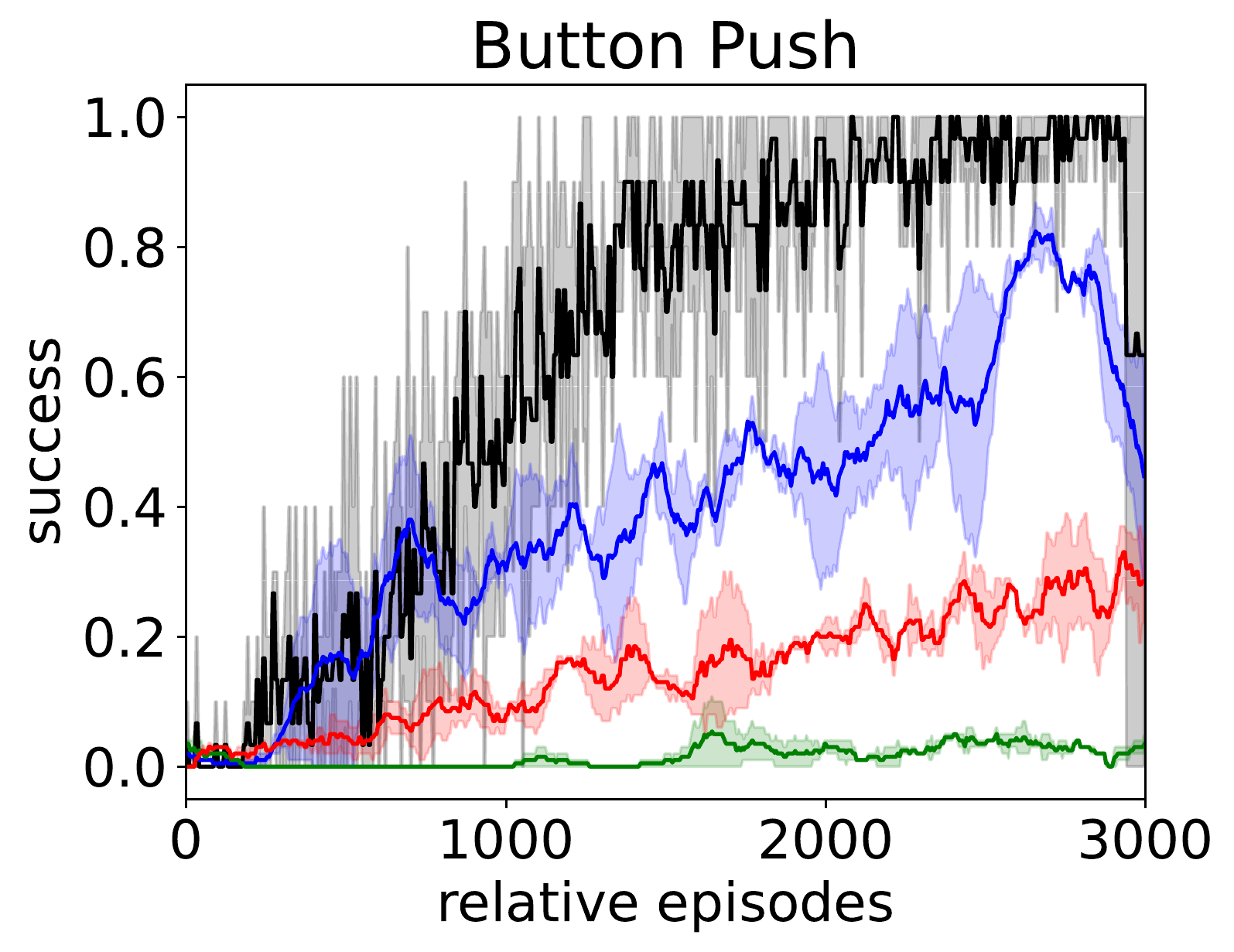}\hfill
    \includegraphics[width=0.20\textwidth]{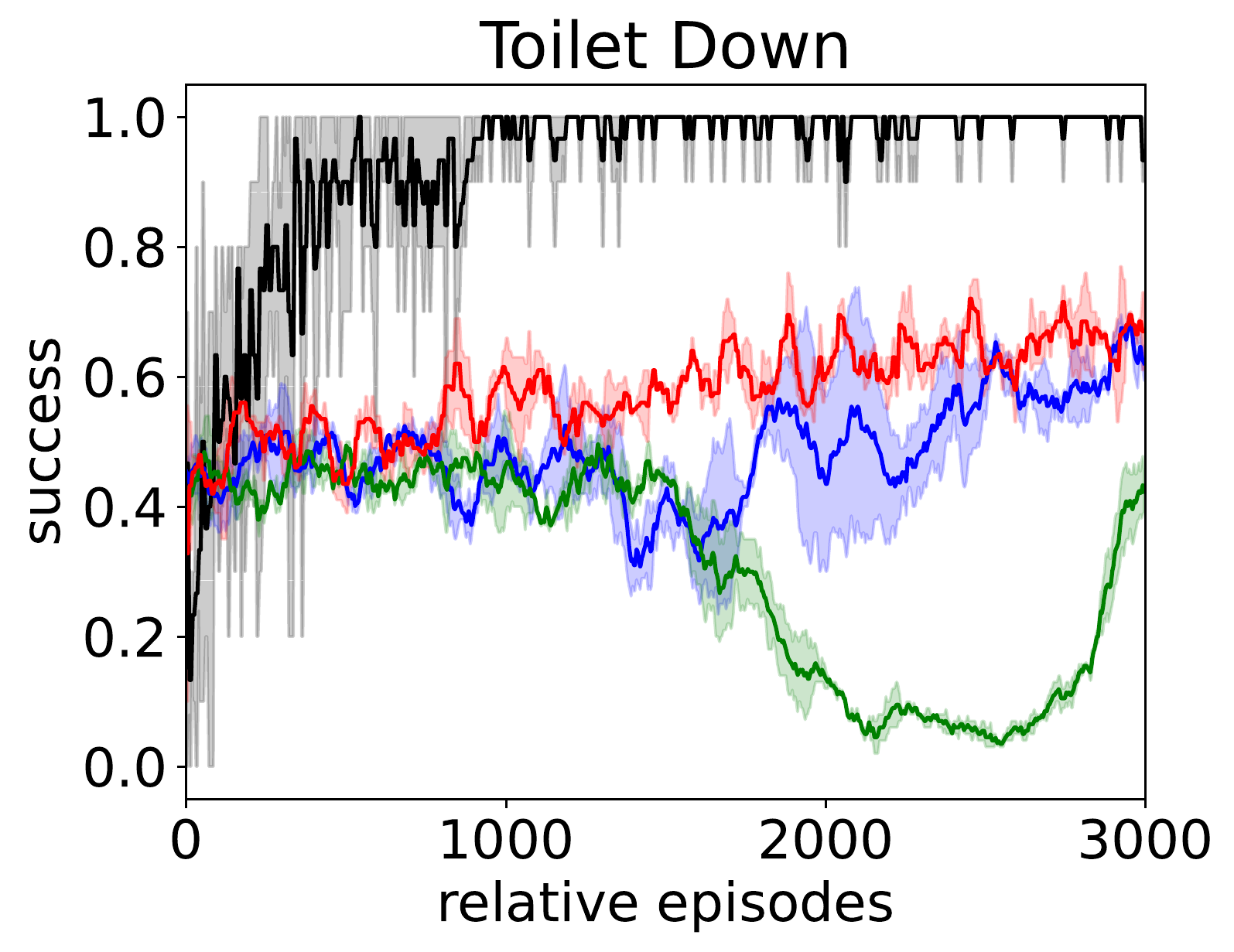}\hfill
    \includegraphics[width=0.20\textwidth]{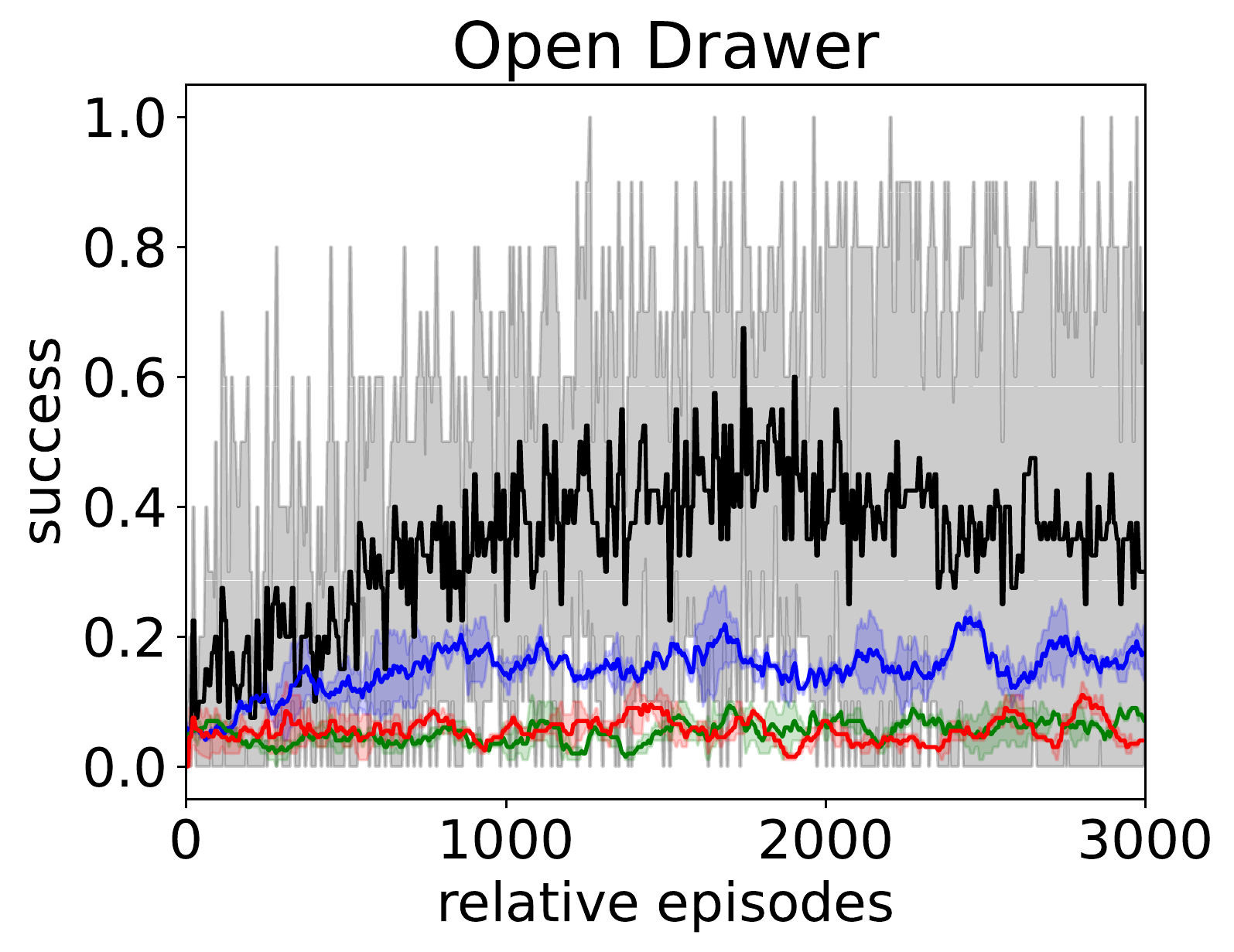}\hfill
    \includegraphics[width=0.20\textwidth]{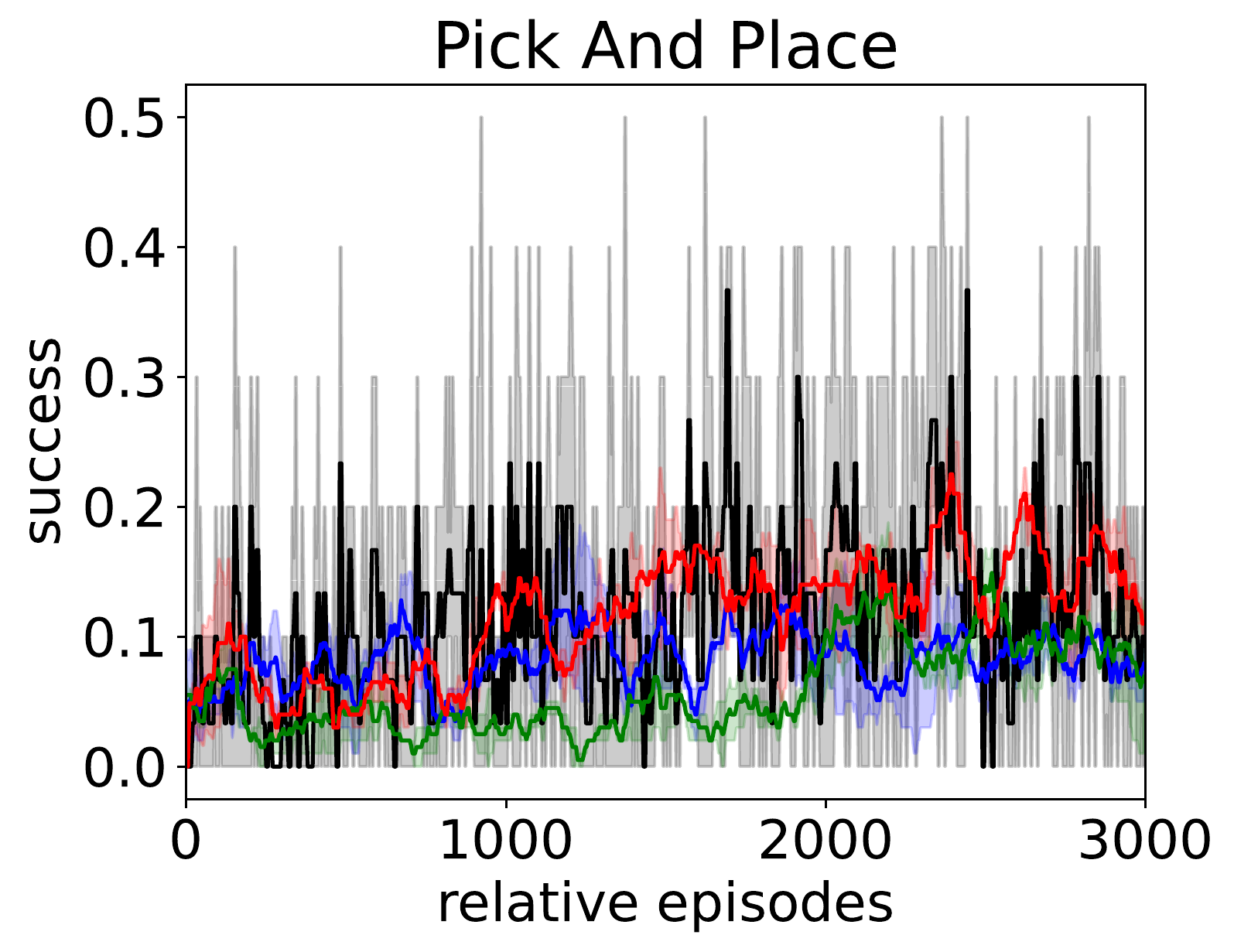}
    \\[\smallskipamount]
    \includegraphics[width=0.20\textwidth]{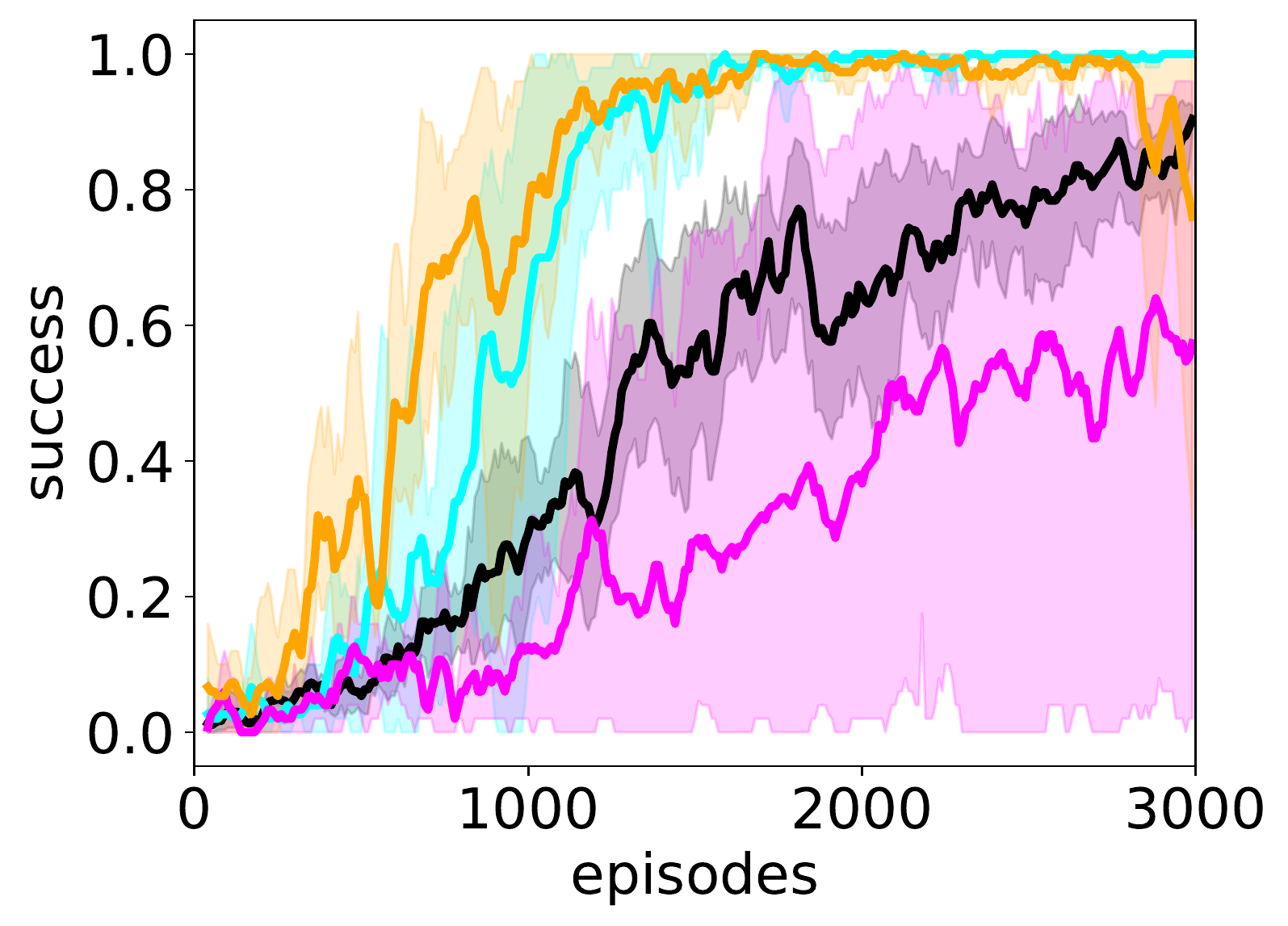}\hfill
    \includegraphics[width=0.20\textwidth]{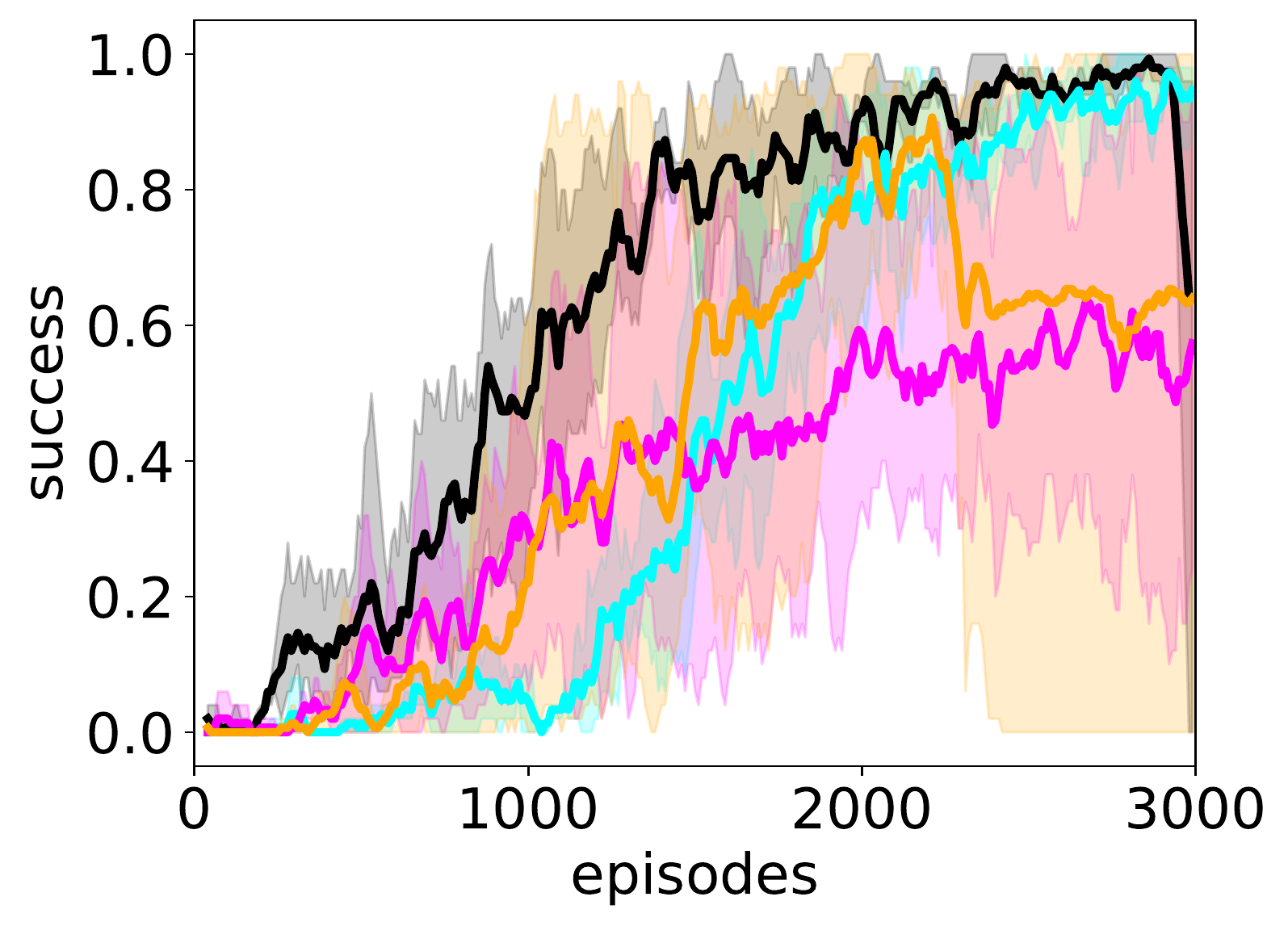}\hfill
    \includegraphics[width=0.20\textwidth]{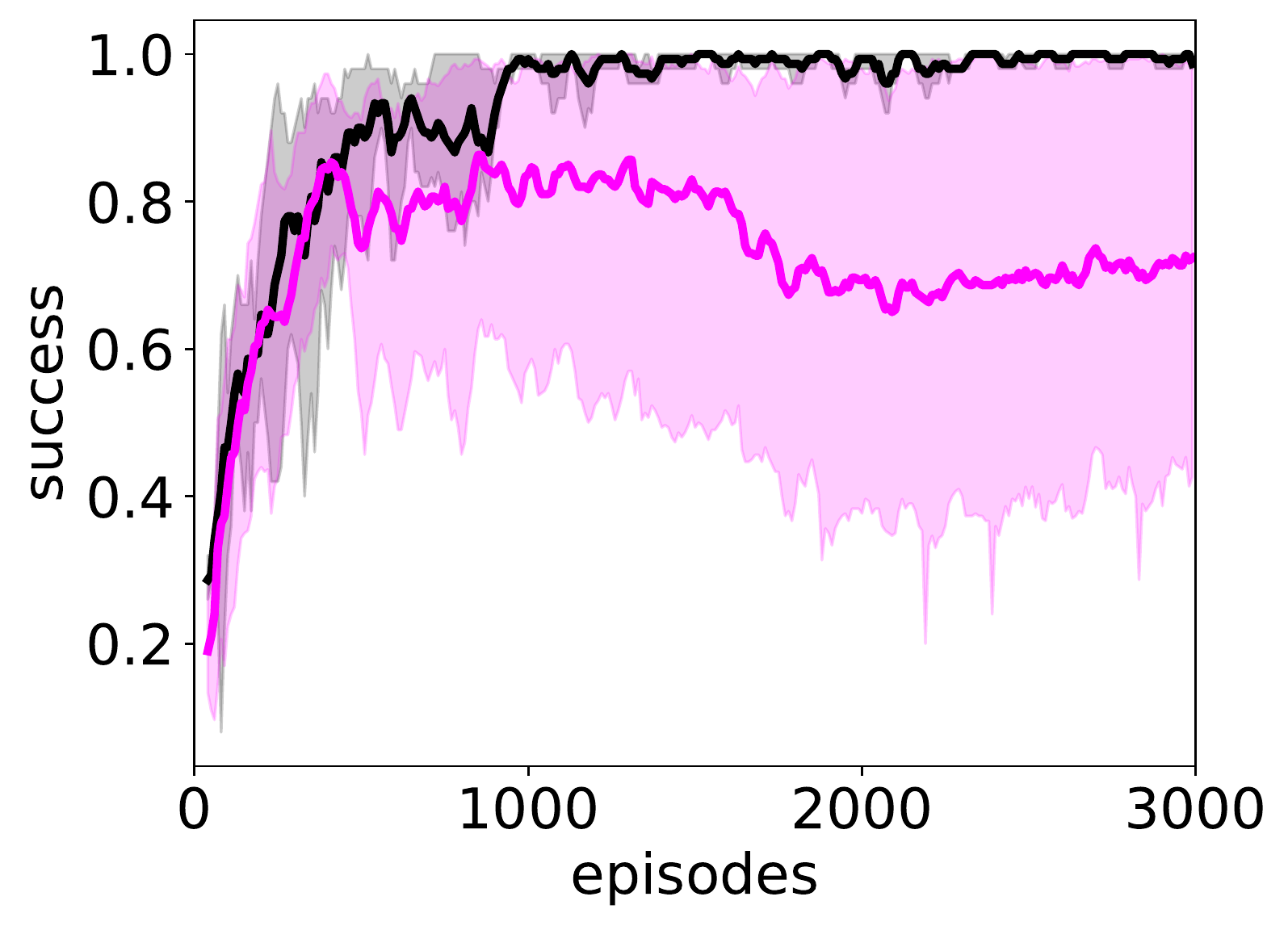}\hfill
    \includegraphics[width=0.20\textwidth]{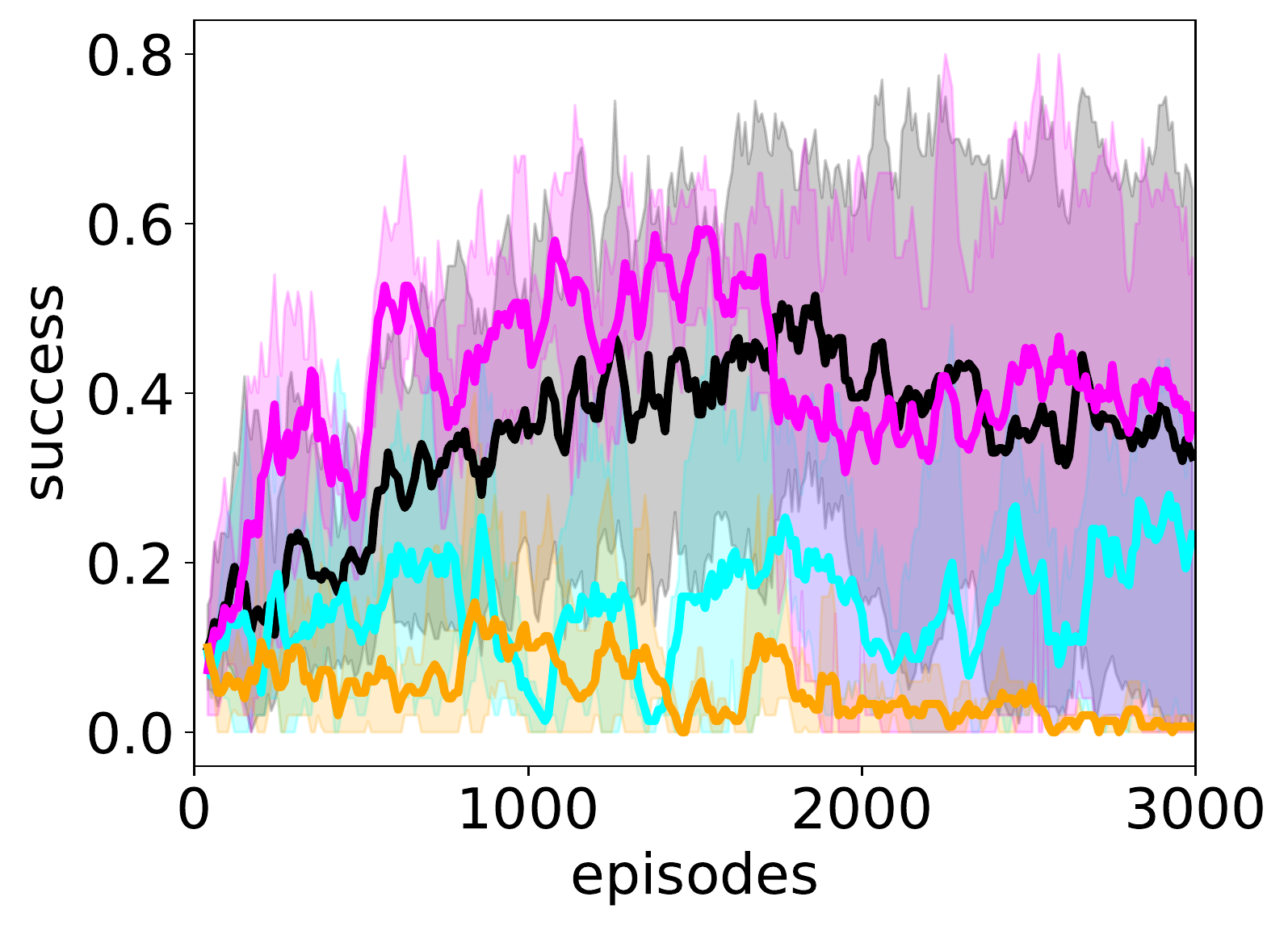}\hfill
    \includegraphics[width=0.20\textwidth]{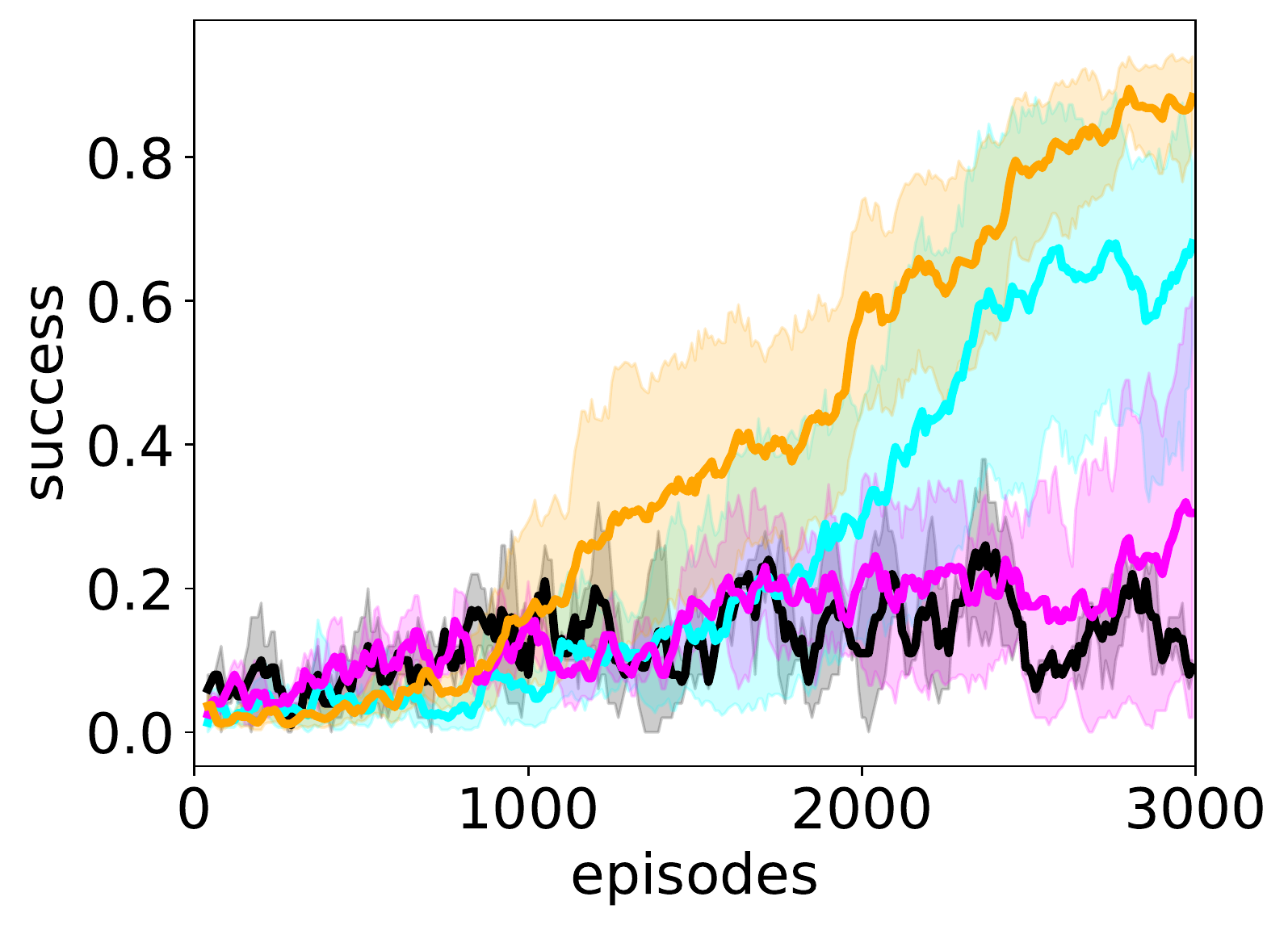}
    \\[\smallskipamount]
    \includegraphics[width=\textwidth]{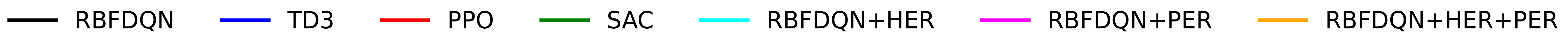}
    %\\[\smallskipamount]
    %\includegraphics[width=\textwidth]{replay_legend.pdf}
    \caption{Policy evaluation success rate for RLBench tasks \textit{Fetch Reach, Button Push, \textit{Toilet Down}, Open Drawer,} and \textit{Pick and Lift}. The upper row compares RBF-DQN with baselines TD3, PPO, and SAC. The bottom row compares RBF-DQN combined with HER, PER, and HER+PER. Data is averaged over a rolling window of size 5. Each episode is 200 steps. For \textit{Fetch Reach}, PPO converged at around 2000 iterations, with 10 update epochs in each iteration and considered as 20000 episodes equally. TD3 converged at around 13000 episodes, and SAC converged at around 12000 episodes. For \textit{Button Push}, PPO converged around at 1300 iterations or 13000 episodes, TD3 converged at around 7000 episodes and SAC converged at 14000 episodes. For \textit{Toilet Down}, PPO converged after 1300 iterations or 13000 episodes, TD3 converged after more than 11000 episodes and SAC needs at least 13000 episodes to converge. For \textit{Open Drawer}, PPO achieved a maximum success rate of 0.30 after 3000 episodes, TD3 achieved a success rate of 0.30 after 5000 episodes and SAC can hardly achieve such a success rate along the process. We run all algorithms for 3 seeds and shade the 95\% confidence interval for each run.}
    \label{fig:my_label}
\end{figure*}
We evaluated RBF-DQN along with state of the art baseline implementations of SAC, TD3, and PPO \cite{stable-baselines3}. All algorithms were tested on five tasks in RLBench \cite{james2019} using a Franka Panda Arm with 8 DoF (7 joints + 1 gripper tip), each with a continuous range of motion. Agents receive a reward of 1 when they completes the task and a reward of 0 for all other time steps. All variations are trained in the joint velocity action space, where actions are represented as an 8 dimensional vector, where each element corresponds to a joint velocity or gripper tip open position. For each task, we used the low dimensional state space provided by RLBench, consisting of information about the robot arm joint velocities, and all objects in the scene. The state space was pruned to reduce its dimensionality and remove irrelevant information. Each agent was trained for 3,000 episodes, where each episode corresponds to a maximum of 200 steps.

Descriptions of the tasks, initialization sequences, and state spaces are described below.

\textbf{Reach} and \textbf{Button Push}: The robot arm is required to move to a target position in the environment (and push a button). The state space is 17 dimensional, representing the joint positions of the arm, position of the end effector tip, and position of the target to reach. Goals for HER on the Fetch Reach and Button Push task are derived from the ending $(x, y, z)$ end-effector position. 

\textbf{Toilet Seat Down}: The robot arm is required to put the lid of a toilet seat down. The state space is 101 dimensional, where the state encompasses information about the gripper joint positions and velocities as well as information about the toilet, like its position, orientation and joint state (how open or closed the lid is). Goals for HER are based on the toilet lid joint. 

\textbf{Pick and Place}: The robot arm is required to pick up a block and move it to a spot in 3D space. This task proved extremely difficult in the sparse reward setting, so we simplified the task by first motion planning to the block, forming and maintaining a grasp throughout the trajectory (locking the 8th element of the action vector to keep the gripper closed). The reduced state space is 51 dimensional, representing the robot joint velocities, block position, and target position in space. Goals for HER are formed from the $(x, y, z)$ position of the end effector. 

\textbf{Open Drawer Task}: The robot arm is required to pull open a drawer. Due to the difficulty of this task in the sparse reward setting, we initialize the gripper at the beginning of each episode to make durative contact with the bottom handle of the drawer, and form a grip. Throughout the trajectory, the robot has full control over its 8 dimensional action space. The reduced state space consisted of 45 dimensions: the joint velocities and gripper state of the robot arm, the waypoint of the bottom drawer, and the prismatic joint of the bottom drawer (loosely representing how open or closed the drawer is). Goals for HER were formed with the drawer's prismatic joint, which increases from 0 as the drawer is opened. 
\section{Discussion and Analysis}
% Re-summarize Results -- need to rewrite based on what our new results look like. What are our main conclusions for: (1) RBF-DQN vs baselines, (2) RBF-DQN vs w/ sample efficient strategies
From the results, we see RBF-DQN under an $\varepsilon$-greedy policy compares favorably to other state-of-the-art baselines under the same conditions. In the five sparse reward RLBench robotic manipulation tasks evaluated, RBF-DQN required 1/3 as many episodes to succeed at each task, which is a significant breakthrough in sample efficiency for robotic manipulation.

% Analysis of Results -- (1) why is RBF-DQN more sample efficient than baselines? (2) Why is HER/PER ineffective on certain tasks, to the point where performance can be detrimental?
%Simon's version
We note that while RBF-DQN was successful, not all of the sampling strategies using (HER, PER, HER+PER) were equally effective on each task: PER may result in unstable learning, and HER may not always be feasible to incorporate. \textit{Fetch Reach} and \textit{Pick and Place} had success under HER and HER+PER, but when using PER only, the training had a tendency to become unstable; for \textit{Button Push}, neither PER, HER or HER+PER outperformed vanilla RBF-DQN; 
for \textit{Open Drawer}, HER did not increase performance, while PER increased learning speed but was unstable. For \textit{Toilet Down}, as a result of their being no intermediate stable goal states (the lid is either up, or falling down due to gravity with slight perturbations), HER is not useful, and PER leads to unstable learning compared to vanilla RBF-DQN.
%original version
%We note that while RBF-DQN was successful, not all of the sampling strategies used ($\varepsilon$-greedy, HER, PER) were equally effective on each task: \textit{Fetch Reach} had success under all learning strategies; \textit{Button Push} succeeded with $\varepsilon$-greedy and HER, but not PER; \textit{Toilet Down} succeeded with $\varepsilon$-greedy and PER, but not HER; \textit{Pick and Place} succeeded with HER and PER, but not $\varepsilon$-greedy; and \textit{Open Drawer} succeeded with $\varepsilon$-greedy and PER, but not HER, and $\varepsilon$-greedy performance decreased at the end.

% Explain HER/PER in environment context
% things we shall cover here: 1. why PER only is not working. 2. why her doesn't help in the button push and open drawer case. 3. 
Differences in the environment may play a role in the success of the sampling strategies in terms of what areas of state space the agent explored. In particular, the stability of trajectory states sampled from the experience buffer (and those which are chosen as hindsight goals) may have an impact on success. \textit{Fetch Reach} and \textit{Button Push} have the property that all states in the state space are stable: in the absence of robotic control, states (of the end effector or the button) do not transition to a fixed point.

For \textit{Toilet Down}, the goal state of the toilet lid is an attractor for lid joint angle due to gravity, so certain perturbations of the lid when it is open can cause the lid to fall to the goal state. Setting hindsight goals for lid angles which naturally fall towards the goal state, with no robot contact on the lid, could be a successful hindsight goal selection strategy. However, since the robot does not need an intelligent policy at the subgoal states due to the attraction dynamics of this task family, hindsight goal selection is not as beneficial as in cases where planning is challenging from the subgoal states.

%For \textit{Toilet Down}, the goal state of the toilet lid is an attractor for lid joint angle due to gravity, so certain perturbations of the lid when it is open can cause the lid to fall to the goal state. Setting hindsight goals for lid angles which naturally fall towards the goal state, with no robot contact on the lid, could be a successful hindsight goal selection strategy. However, this ignores that the hindsight goal is out of the desired goal distribution.

Additionally, the mapping $\phi$ from state space to goal space is critical: certain tasks can only be completed if the robot successfully maintains durative contact with the object throughout the trajectory. Therefore, certain tasks require hindsight goals to be created out of states that maintain durative contact. \textit{Pick and Place} and \textit{Open Drawer} states are stable, but only as long as the gripper maintains contact with the object or handle. Therefore, in \textit{Pick and Place}, we opted to always ensure the gripper remained closed. In contrast, for \textit{Open Drawer}, we performed only grasp initiation, but subsequently allowed the robot arm full control over its DoF, implying that it could potentially release its grip on the drawer. We observe that due to these differences, HER on \textit{Pick and Place} was more effective than HER on \textit{Open Drawer}, since in \textit{Open Drawer}, there is a very low probability that the gripper remains closed throughout the trajectory. 

In certain experiments such as \textit{Button Push}, \textit{Fetch Reach}, and \textit{Open Drawer}, HER + PER resulted in performance of the agent collapsing near the end of training. It is possible that the bias introduced by Priority Experience Replay is significant enough to destabalize convergence at the end of training, and that the weighted importance sampling ratios in PER would benefit from an annealing schedule that reduced the weights over time. This is especially problematic for value-based approaches like RBF-DQN which tend to be less stable during training than policy-gradient methods like TD3, PPO and SAC since they optimize for low Bellman error rather than directly improving the expected returns of the policy. Future work will investigate approaches for mitigating the destabilization issues introduced by biased replay buffer sampling techniques. 
%erroneously prioritizing outlier samples (which occur infrequently and have a high TD error). The net result could be an overfitting effect, whereby the agent overfits to outlier samples, straying away from a more generalizeable policy. 
%In certain experiments such as \textit{Button Push}, \textit{Fetch Reach}, and \textit{Open Drawer}, HER + PER resulted in performance of the agent collapsing near the end of training. It is possible that Priority Experience Replay is erroneously prioritizing outlier samples (which occur infrequently and have a high TD error). The net result could be an overfitting effect, whereby the agent overfits to outlier samples, straying away from a more generalizeable policy. 

% Relation of Results to Field
Even without common sample-efficiency improvements, we have demonstrated that RBF-DQN is relatively more sample efficient than current state of the art baselines, and is able to perform better or comparably on multiple robot manipulation tasks. We attribute the success of RBF-DQN on sparse reward, continuous state \& action manipulation tasks to the action maximization and function approximation properties of RBFs, which guarantee the location of the max centroids to approximately correspond with the max Q value at a given state, within an $\mathcal{O}(e^{-\beta})$ error. This property is extremely powerful, allowing action maximization to be achieved over simple centroid search (of which there are finitely many), suggesting why RBF-DQN performs efficiently.
% What did this problem solve?

% Given this result, what can now be done?
Our results provide strong motivations for incorporating RBF-DQN as a sample-efficient value-based method in the domain of robotic manipulation.
% Future Directions / Limitations (Or use the conclusion for this)
It seems promising that RBF functions can be leveraged to improve sample complexity for robotic manipulation tasks with both on-line and off-line RL methods. It would be interesting to see how RBF-DQN performs in higher dimensional state representations, and how other sampling methods or goal generation methods could be utilized to improve sample efficiency.
\section{Conclusion}
\label{sec:conclusion}
We have experimentally seen that RBF-DQN is comparable or better at common robotic manipulation tasks to PPO, TD3, and SAC. Especially when paired with HER and PER, RBF-DQN is a powerful value based model for off-policy continuous action space robotic manipulation. 

In the future, we hope to experiment with using RBF-DQN on vision based state input (depth images, and point clouds), incorporating sample efficiency algorithms like CURL \cite{curl} and RAD \cite{rad}, as well as adapting HER to work with image based state input. Furthermore, we are working to improve the stability and over-estimation tendencies of RBF-DQN by exploring the potential of incorporating dueling \cite{duelingDQN} and double \cite{doubleDQN} DQN techniques into RBF-DQN. 
\section*{Acknowledgments}

%% Use plainnat to work nicely with natbib. 

\bibliographystyle{plainnat}
\bibliography{main}

\end{document}